\documentclass[11pt]{article}
\usepackage[a4paper,margin=0.97in]{geometry}
\usepackage{authblk}
\usepackage[T1]{fontenc}
\usepackage[utf8]{inputenc}
\usepackage{lmodern}
\usepackage{amsmath,amssymb,amsthm,bm}
\usepackage{booktabs,tabularx,array,multirow}
\usepackage{graphicx}
\usepackage{etoolbox}
\usepackage[numbers,sort&compress]{natbib}
\usepackage[colorlinks=true,linkcolor=blue,citecolor=blue,urlcolor=blue]{hyperref}
\newcommand{\Prob}{\Pr}

\DeclareMathOperator{\logit}{logit}

\newtheorem{definition}{Definition}

\AtBeginEnvironment{table}{\small}

\title{Selective Verification of Transferred Evidence-Interaction Explanations in Retrieval-Augmented Language Models}
\date{}

\author[1]{Sidra Nasir}
\author[2]{Rizwan Ahmed Khan\thanks{Corresponding author: RizwanKhan@iba.edu.pk}}

\affil[1]{Dipartimento di Informatica, Universit\`a di Verona, Verona, Italy}
\affil[2]{Department of Computer Science, School of Mathematics and Computer Science (SMCS),
Institute of Business Administration (IBA), Karachi, Pakistan}
\begin{document}
\maketitle

\begin{abstract}
Passage-deletion effects and interactions in retrieval-augmented generation (RAG) can differ across language-model readers, making direct transfer of evidence-use explanations unreliable. We formulate transfer as selective verification of intervention-defined claims under a limited target-response budget. For each question, four fixed passages define 16 contexts and a response surface from which we compute four deletion effects and six pairwise interactions. A coefficient-wise mixture prior borrows source-reader information while retaining a diffuse target-specific component. Sequential Monte Carlo incorporates target observations selected by a claim-directed Bayesian investigator or a recurrent proximal policy optimization (PPO) investigator. Across five model families and three multi-hop question-answering datasets, Bayesian verification attains mean noise-corrected contrast error of 0.025 at 32 target responses, versus 0.040 for target-only inference (paired difference $-0.0150$, Holm-adjusted $p<0.001$). In an independent Qwen3-235B predictor audit, Bayesian-verified explanations yield a noise-corrected Brier score of 0.078, compared with 0.100 without explanations. PPO attains error of 0.026, without a statistically supported advantage over Bayesian acquisition. Claim-level correction, negative-transfer controls, and independent reference measurements delimit the reliability of transferred explanations.
\end{abstract}


\section{Introduction}
\label{sec:introduction}
Retrieval-augmented generation (RAG) conditions language-model answers on retrieved passages \citep{lewis2020rag}. Removing a passage measures its effect on answer correctness under a defined intervention; this quantity is not interchangeable with retrieval relevance, generated citations, or an explanation of internal model computation \citep{cohenwang2024contextcite,liu2025attribot,wallat2025correctness}. Passage utility can change when the reader changes, even if the question and retrieval context remain fixed \citep{zhang2026utility,zhou2026preference,wu2024clasheval}.

Multi-hop evidence can have conditional effects: the contribution of one passage may depend on the presence of another \citep{kang2025spex,jain2026cuer}. An explanation transferred between readers must therefore support separate decisions about passage-deletion effects and interactions. Neither copying a source explanation nor estimating every effect from target observations alone addresses the decision of which borrowed claims require verification under a limited target-response budget.

We model each reader through correctness probabilities under the 16 subsets of four fixed passages. Ten linear contrasts encode four full-context deletion effects and six conditional pairwise interactions. The transfer model assigns independent source-mixture priors to coefficients of a complete Walsh-logit representation and includes a diffuse target-specific component. Sequential target observations update the prior through a particle posterior; a Bayesian investigator selects interventions by expected reduction in claim-decision risk, while a recurrent PPO policy provides an alternative acquisition strategy.

The contributions are: (i) a claim-level cross-reader transfer formulation with explicit confirmation, withdrawal, and reversal decisions; (ii) coefficient-wise partial borrowing with target-specific fallback and sequential posterior updating; and (iii) an evaluation separating continuous contrast estimation, claim reliability, source mismatch, and independent counterfactual prediction. Source-mixture modeling, sequential Monte Carlo, and reinforcement learning are existing techniques; the methodological contribution concerns their specified use for selective verification of transferred evidence-interaction claims. All claims refer to observed correctness under context interventions rather than latent model mechanisms.

\section{Related work}
\label{sec:related}
Context attribution estimates passage effects from controlled interventions, while counterfactual evaluation measures their predictive utility. Reader-dependent passage use introduces a transfer problem when an effect measured for one model is applied to another. Related methods address attribution, interaction estimation, and sequential transfer under different observation settings.
\subsection{Attribution and evidence interactions}
ContextCite and AttriBoT estimate context dependence using ablations and approximations to their effects \citep{cohenwang2024contextcite,liu2025attribot}. Citation correctness and causal reliance are distinct \citep{wallat2025correctness}. SPEX, ProxySPEX, and Shapley-based methods also consider interactions between evidence components \citep{kang2025spex,butler2025proxyspex,fumagalli2024kernelshapiq}. Our task differs in asking whether a previously estimated passage effect or interaction continues to hold for another reader.

\subsection{Reader dependence and cross-model transfer}
The utility and sign of evidence effects can differ across readers \citep{zhang2026utility,zhou2026preference,wu2024clasheval,liu2024lost}. ProEval and AP-LSE address related cross-model evaluation and sequential-prior transfer problems \citep{huang2026proeval,ngo2025aplse}. We include task-specific adaptations of these methods as comparators, without treating them as reproductions of the original procedures. The unit of transfer in our formulation is an individual deletion or interaction claim that may also be withheld.

\subsection{Verification and counterfactual simulatability}
Counterfactual simulatability measures whether an explanation helps predict reader behavior after an intervention \citep{chen2024simulatability,lyu2024faithful,karvonen2026chive}. Bayesian and learned experimental design provide methods for choosing informative observations sequentially \citep{rainforth2024modern,filstroff2024targeted,foster2021dad,hedman2025stepdad}. We apply these principles to select target-reader interventions and evaluate transferred claims against independent responses.
\section{Problem formulation}
\label{sec:problem}
A fixed retrieved context defines 16 interventions on four passages. The correctness probabilities under these interventions determine deletion and pairwise interaction contrasts. Source and target response histories determine the information available for transfer and subsequent claim verification.

\subsection{Response surfaces and evidence-use claims}

Let $q$ be a question with accepted answers $A_q$, and let $D_q=(d_1,d_2,d_3,d_4)$ be four retrieved passages in a frozen order. A subset $S\subseteq D_q$ defines a context intervention: the model receives the question together with exactly the passages in $S$, in their frozen relative order. For a model endpoint $m$ under fixed decoding configuration $g$, let $Z$ denote a freshly sampled response and define
\begin{equation}
\begin{aligned}
Y_{q,m,S}&=\mathbf 1\{\operatorname{normalise}(Z)\in\operatorname{normalise}(A_q)\},\\
p_{q,m}(S)&=\Prob\left(Y_{q,m,S}=1\mid q,S,g\right).
\end{aligned}
\label{eq:observation}
\end{equation}
Under stochastic decoding, $p_{q,m}(S)$ is the probability of a correct response, not the outcome of an individual request. The 16 probabilities constitute the reader's \emph{response surface} for a fixed endpoint and decoding configuration. Estimation assumes a stationary endpoint and independent requests.

From the surface we derive ten contrasts. The deletion effect of passage $i$ is
\begin{equation}
\delta_i=p(D_q)-p(D_q\setminus\{d_i\}),
\label{eq:deletion}
\end{equation}
and the normalised conditional interaction of passages $i$ and $j$ is
\begin{equation}
\gamma_{ij}=\tfrac12\bigl[p(D_q)-p(D_q\setminus\{d_i\})-p(D_q\setminus\{d_j\})+p(D_q\setminus\{d_i,d_j\})\bigr].
\label{eq:interaction}
\end{equation}
Both contrasts lie in $[-1,1]$. The identity $2\gamma_{ij}=[p(D_q)-p(D_q\setminus\{d_i\})]-[p(D_q\setminus\{d_j\})-p(D_q\setminus\{d_i,d_j\})]$ shows that positive $\gamma_{ij}$ corresponds to a greater marginal benefit from retaining passage $i$ when passage $j$ is present. Negative values correspond to subadditive effects. These contrasts describe changes in response correctness, not internal computations. Let $e=Cp$, where $C\in\mathbb{R}^{10\times16}$.

\begin{definition}[Evidence-use explanation]
For a prespecified materiality threshold $\epsilon>0$, the claim associated with contrast $e_k$ is its region $r_k\in\{-,0,+\}$, where $r_k={-}$ if $e_k<-\epsilon$, $r_k={+}$ if $e_k>\epsilon$, and $r_k=0$ otherwise. The evidence-use explanation of model $m$ for question $q$ is the tuple of ten claims $(r_1,\dots,r_{10})$, together with the full-context correctness rate $p(D_q)$.
\end{definition}

We use $\epsilon=0.20$ to assign negative, neutral, and positive claim regions. The narrower interval $[-0.10,0.10]$ defines the practical-equivalence criterion used by target-only certification; the neutral claim region $[-0.20,0.20]$ does not itself certify equivalence. A positive deletion contrast denotes a beneficial passage under the full-context comparison, whereas a negative contrast denotes a harmful one. Positive interactions represent conditional complementarity. We report a material-region claim as confident when its posterior region probability exceeds $\kappa=0.80$.

A claim is conditional on the specified passages and context intervention. It neither identifies a unique cause of the generated answer nor provides an explanation of the reader's internal mechanism.

\subsection{Explanation transfer}

Consider $M$ endpoints. When a target endpoint $t$ is introduced, histories from the $M-1$ source endpoints are available: for each source $s$ and subset $S$, $k_{s,S}$ correct responses out of $n_{s,S}$. The investigator may then request at most $B$ responses from $t$, one subset at a time, observing a binary outcome after each request.

\begin{definition}[Explanation transfer problem]
Given the source histories $H=\{(k_{s,S},n_{s,S}):s\neq t\}$ and a budget of $B$ sequential target observations, produce region probabilities for the ten claims of target $t$ that are accurate with respect to the target's latent surface $p_t$, with conclusions that remain calibrated when source claims do not hold for $t$.
\end{definition}

A \emph{borrowed} explanation uses only source histories ($B=0$). A \emph{verified} explanation also incorporates up to $B$ target responses. For an initially confident claim, the target posterior may retain the same region (confirmed), assign confidence to another region (overturned), or no longer support a confident region (withdrawn). Claims that become confident only after verification are designated new. The objective is to correct unsupported transferred claims without unnecessarily changing valid ones.

The warm setting permits source histories for the same question. The new-question setting excludes these histories and predicts prior parameters from question and passage features learned on source data. Target data are excluded from training and hyperparameter selection.

\section{Selective transfer of evidence-interaction explanations}
\label{sec:method}

\begin{figure}[tbp]
    \centering
    \includegraphics[width=\linewidth]{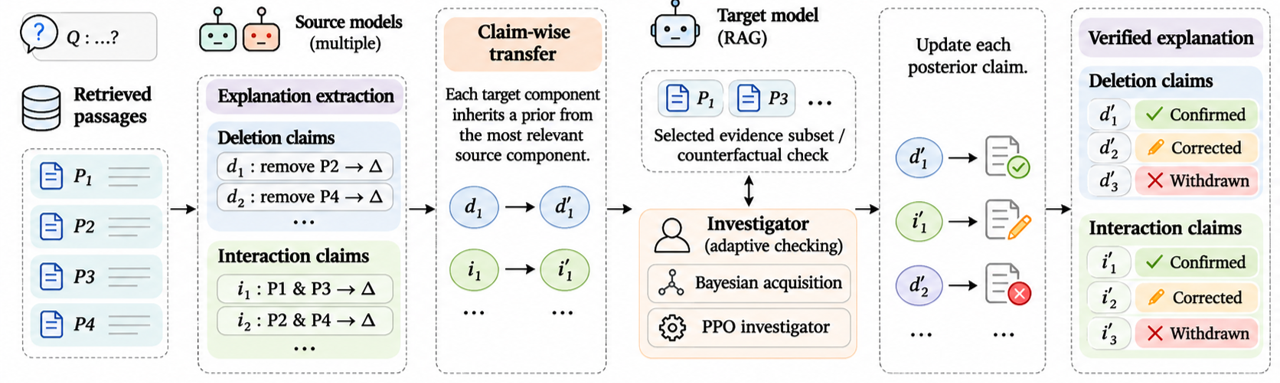}
    \caption{Selective verification procedure for transferring deletion and interaction claims between retrieval-augmented language-model readers.}
    \label{fig:methodology}
\end{figure}

Source response histories determine a prior over the target response surface. Sequential target interventions update this distribution, and posterior region probabilities determine the reported deletion and interaction claims. Figure~\ref{fig:methodology} depicts the verification procedure.

\subsection{A complete logit representation of the response surface}

Encode a subset by its indicator vector $x(S)\in\{0,1\}^4$. For each coefficient index $T\subseteq\{1,2,3,4\}$ let
\begin{equation}
\phi_T(S)=\prod_{i\in T}\bigl(2x_i(S)-1\bigr),\qquad
\logit p(S)=\sum_{T\subseteq\{1,2,3,4\}}\theta_T\,\phi_T(S).
\label{eq:basis}
\end{equation}
The $16\times16$ Walsh matrix $W_{S,T}=\phi_T(S)$ satisfies $W^\top W=16I$, giving $\theta=W^\top\logit(p)/16$. This basis provides a complete parameterization of the logit response surface. The reported explanation contrasts are nevertheless computed from probabilities, not directly from the basis coefficients.

For source $s$ with $k_{s,S}$ correct responses out of $n_{s,S}$, independent Jeffreys posteriors
\begin{equation}
p_{s,S}\mid H_s\sim\operatorname{Beta}\left(k_{s,S}+\tfrac12,\;n_{s,S}-k_{s,S}+\tfrac12\right)
\label{eq:sourcebeta}
\end{equation}
are sampled 512 times. Each sample is transformed to the logit basis to estimate source-specific coefficient means $\mu_{s,T}$ and standard deviations $\sigma_{s,T}$, retaining sampling uncertainty from finite source histories.

\subsection{A coefficient-wise mixture prior for partial transfer}

For each coefficient, the prior over the target is
\begin{equation}
\begin{aligned}
\pi_T(\theta_T) &= \rho\,\mathcal N(\theta_T;0,\tau^2)
 + \frac{1-\rho}{M_s}\sum_{s=1}^{M_s}
 \mathcal N(\theta_T;\mu_{s,T},v_{s,T}^2),\\
v_{s,T} &= \max(\lambda\sigma_{s,T},0.05).
\end{aligned}
\label{eq:mixture}
\end{equation}
The joint prior factorizes as $\pi(\theta)=\prod_T\pi_T(\theta_T)$, with $M_s=4$ source readers, $\rho=0.25$, $\lambda=1$, and $\tau=3$. The source mixture permits coefficient-specific borrowing, while the diffuse component weighted by $\rho$ permits deviations from every observed source. Posterior samples are mapped back to response probabilities before computing deletion and interaction contrasts. Whole-source and no-diffuse-component variants test the importance of these prior choices.

\subsection{Sequential Monte Carlo inference}

Particles $\theta^{(r)}$ with weights $w^{(r)}$, $r=1,\dots,512$, approximate the posterior. After observing outcome $Y_t$ for subset $S_t$, weights are updated by the Bernoulli likelihood,
\begin{equation}
w^{(r)}\leftarrow w^{(r)}\,p_{\theta^{(r)}}(S_t)^{Y_t}\bigl[1-p_{\theta^{(r)}}(S_t)\bigr]^{1-Y_t},
\label{eq:weights}
\end{equation}
After each response, the weights are normalized. If the effective sample size falls below 256, the algorithm resamples and applies five Metropolis rejuvenation steps with proposal scale 0.2 \citep{delmoral2006smc}. Weighted particles then provide posterior means, intervals, and claim-region probabilities. Additional development runs examine sensitivity to particle count.

\subsection{Bayesian investigator: claim-directed acquisition}

Let $R(\pi_t)=\sum_{k=1}^{10}\bigl[1-\max_c \Prob(r_k=c\mid H_t)\bigr]$ be the posterior Bayes risk of the ten region decisions. For each subset $S$ whose observation count is below the per-subset acquisition cap of eight, the investigator evaluates the expected reduction in claim-decision risk,

\begin{equation}
a(S)=
\frac{
R(\pi_t)-
\sum_{y\in\{0,1\}}
\Prob(Y=y\mid S,H_t)\,
R\bigl(\pi_t^{S,y}\bigr)
}{
c(S)
}.
\label{eq:acquisition}
\end{equation}

The hypothetical posterior $\pi_t^{S,y}$ is approximated by reweighting the current particles. All candidate requests have unit cost and repeated observations of a subset are permitted until its eight-response acquisition inventory is exhausted. Deterministic tie-breaking resolves equal scores. Every fifth request uses an eligible least-observed subset instead of the claim-risk maximizer; the coverage rule is evaluated by ablation. Unlike generic information gain, acquisition directly targets uncertainty in the reported claim decisions \citep{filstroff2024targeted,rainforth2024modern}.

\subsection{Learned investigator: a recurrent PPO policy}

One-step Bayesian acquisition evaluates the immediate expected reduction in claim risk. To investigate whether a learned policy can use longer observation histories, we train a recurrent investigator with a 128-unit gated recurrent unit (GRU). Its inputs include posterior summaries, mixture responsibilities, observation counts, and previous actions and outcomes. It outputs a masked probability distribution over the 16 subsets and an estimated state value \citep{foster2021dad,blau2022optimizing}.

The policy is trained using PPO with generalized advantage estimation and a terminal reward equal to the negative contrast mean squared error \citep{schulman2017ppo,schulman2016gae}. The training configuration uses 2,048 rollout steps, 256-step minibatches, four epochs per update, and a 128-unit recurrent state; learning rates and entropy coefficients are selected from prespecified development grids. Training episodes are constructed from source readers: one source is treated as a pseudo-target while the others supply its prior. Held-out target readers and evaluation questions are not used for policy training. Bayesian and PPO investigators use the same particle-based estimator and target-response budget. Only context interventions are selected; the policy does not modify reader parameters.

\subsection{Claim-level posterior decisions and uncertainty}

For each claim, the verifier reports a posterior contrast estimate, interval, region probabilities, and transfer status. We separately compute conservative simultaneous Clopper--Pearson intervals using target responses only \citep{clopper1934}, with per-comparison level

\begin{equation}
\alpha' =
\frac{\alpha}
{2M \cdot 16 \cdot B_{\max} \cdot P}
\label{eq:cp}
\end{equation}

In this expression, $M=5$ is the number of endpoints, $B_{\max}=8$ is the maximum acquisition replications per subset, and $P=5$ is the number of acquisition permutations. For each subset, Clopper--Pearson endpoints provide a Bernoulli interval at the adjusted tail level. The lower contrast bound sums lower probability endpoints for positive entries of $C$ and upper endpoints for negative entries; the upper contrast bound reverses these choices. A claim receives target-only certification if its propagated contrast interval lies above $\epsilon$, below $-\epsilon$, or entirely inside the practical-equivalence interval $[-0.10,0.10]$. The simultaneous construction covers the stated endpoints, subsets, observation prefixes, and acquisition permutations within a question under stationary independent Bernoulli response streams; it is not a simultaneous guarantee across questions. Source-informed posterior confidence and target-only interval certification are reported separately.

\begin{figure}[tbp]
    \centering
    \includegraphics[scale=0.27]{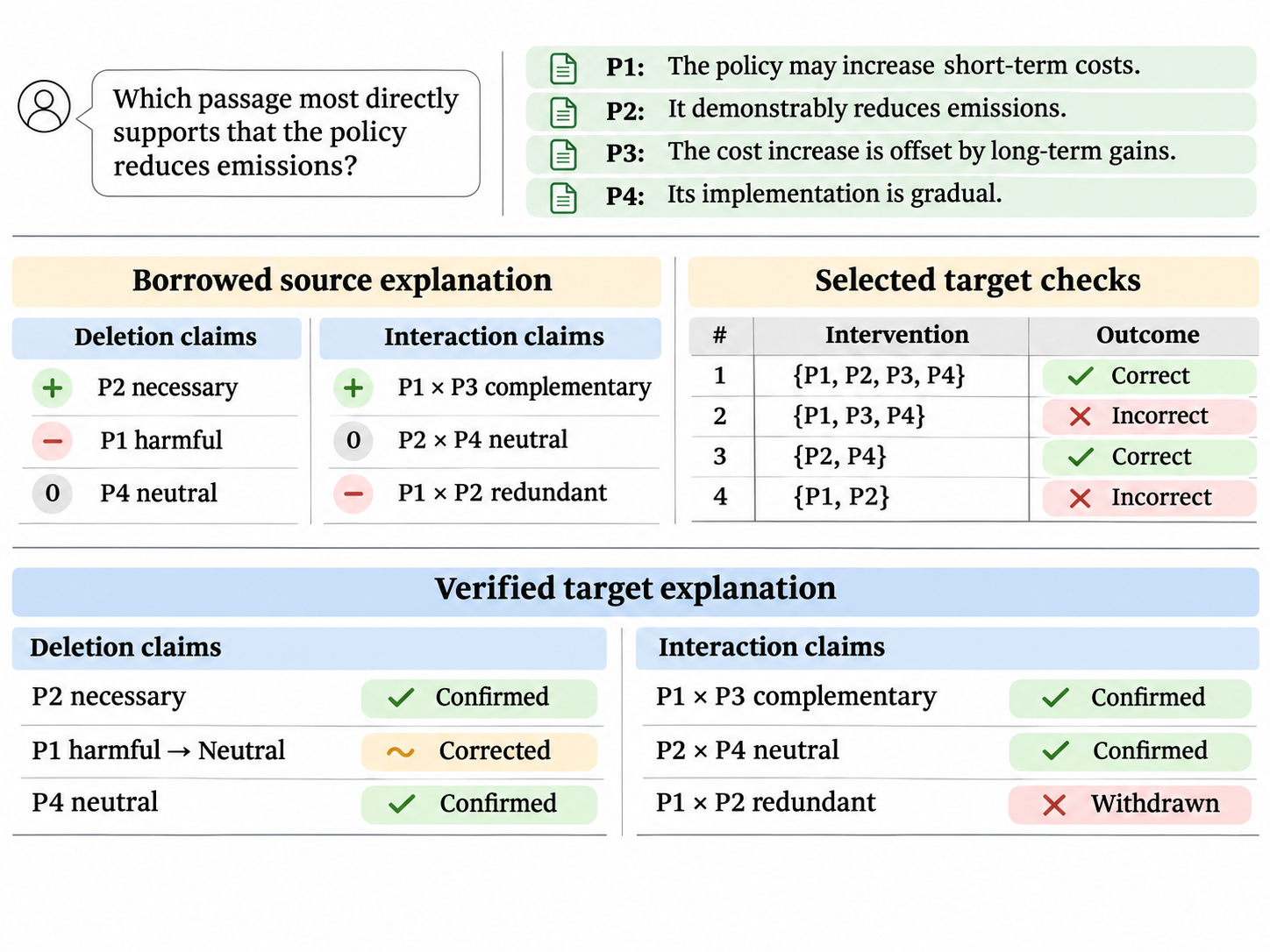}
    \caption{Example of target checks used to confirm, overturn, or withdraw borrowed deletion and interaction claims. Values are illustrative.}
    \label{fig:worked-example}
\end{figure}

Figure~\ref{fig:worked-example} provides a schematic example of source claims, selected target checks, and the resulting claim decisions. Its values are illustrative and are not included in the experimental results.

\section{Experimental evaluation}
\label{sec:experiments}
The evaluation uses separate source histories, sequential target observations, and independent target-reference responses. Target-reference outcomes contribute to scoring but remain unavailable during intervention selection. Independent reference scoring uses $R=16$ responses per subset, distinct from the eight-response maximum for adaptive acquisition from any one subset.

\subsection{Datasets, readers, and evidence interventions}
We evaluate on HotpotQA (distractor), MuSiQue-Ans, and 2WikiMultihopQA \citep{yang2018hotpot,trivedi2022musique,ho2020twowiki}. Each dataset contributes 100 training, 50 development, and 300 held-out questions. Four passages are selected for each question before any model response is observed, using reciprocal-rank fusion of BM25 and dense-retrieval scores \citep{cormack2009rrf,robertson2009bm25,reimers2019sbert}. Passage selection and order are then fixed across readers, including for questions whose selected evidence is insufficient to answer correctly.

Each of the five model families is evaluated as the target, with the remaining four serving as sources (Table~\ref{tab:setting}). All 16 passage subsets are tested using a fixed decoding protocol. Correctness is measured by normalized, alias-aware exact match. The documented decoding configuration uses temperature 0.7, top-$p$ 1.0, and a maximum of 2,048 new tokens, with the same configuration applied to compared reader responses. The source bank contributes four responses per source reader and subset. Source-history, adaptive-acquisition, and reference observations belong to separate banks.

\begin{table}[t]
\centering
\caption{Evaluation setting. Each model is evaluated as a target; the remaining models are potential sources. The table identifies the comparison population and intervention design.}
\label{tab:setting}

\begin{tabularx}{\linewidth}{@{}lX@{}}
\toprule
Dimension & Setting \\
\midrule
Datasets & HotpotQA; MuSiQue-Ans; 2WikiMultihopQA \\
Questions & 100 training, 50 development, 300 evaluation per dataset \\
Evidence intervention & Four fixed passages; all sixteen subsets \\
Model families & Llama-3.1-8B-Instruct; Phi-4; Gemma-3-27B-IT; Qwen3-235B-A22B-\allowbreak Instruct-2507; gpt-oss-20b \\
Explanation targets & Four deletion contrasts and six pairwise interactions \\
Reference outcomes & Independently sampled correctness observations \\
\bottomrule
\end{tabularx}
\end{table}

\subsection{Independent reference observations and comparison methods}
The protocol separates source responses, sequential target responses, and independent target-reference responses. Source observations determine the transfer prior; selected target observations update it; independent reference observations estimate intervention contrasts for scoring. Acquisition cannot access reference outcomes. All streams use the same decoding and answer-normalization protocol. Repeated responses to an identical subset are retained because correctness is a stochastic binary outcome.

We evaluate Bayesian selective transfer and recurrent PPO against borrowing without target verification, target-only inference, and binary-contrast adaptations of AP-LSE and ProEval \citep{ngo2025aplse,huang2026proeval}. The adaptations are task-specific baselines, not exact reproductions of the cited methods. The primary budget is $B=32$ target responses. This equals two observations per subset on average under balanced allocation, although adaptive investigators can distribute their requests unevenly. Primary results and most stratified analyses use this budget with matched target access. Other budgets were considered, but complete trajectories are unavailable; the results therefore do not establish performance across budgets or an optimal budget. The independent target-reference sample contains $R=16$ responses per subset and is disjoint from the observations used for sequential acquisition.

\subsection{Transfer agreement and estimation accuracy}
We measure region agreement for each directed source--target pair and compare it with same-reader and shuffled-question controls. Deletion and interaction claims are evaluated separately.

Let $\bar Y_S$ denote the correctness proportion across $R=16$ independent reference responses for subset $S$, and let $\widehat e^{\mathrm{ref}}=C\bar Y$. The noise-corrected squared error is
\begin{equation}
\widehat L_k=(\widehat e_k-\widehat e^{\mathrm{ref}}_k)^2-
\sum_S C_{kS}^2\frac{\bar Y_S(1-\bar Y_S)}{R-1}
\label{eq:loss}
\end{equation}
For independently collected Bernoulli reference outcomes and an estimate fixed without reference access, the subtraction removes the reference sampling variance and $\widehat L_k$ is unbiased for squared contrast error. The correction can produce negative individual corrected losses; aggregation does not clip such values. We average the corrected errors across ten claims and report deletion and interaction results separately. Claim-region accuracy, confidence precision, and calibration assess other aspects of estimation quality.

\subsection{Verification reliability and methodological ablations}
Claim verification is evaluated separately from contrast regression. Among initially confident borrowed claims, we measure how often an incorrect claim is overturned or withdrawn and how often a correct claim is changed. These rates have different denominators. We also report post-verification region accuracy, confidence coverage, and conditional risk. Conservative target-only intervals provide a distinct certification criterion. Independent reference uncertainty is incorporated when identifying correct and incorrect regions.

Ablations test coefficient-wise versus whole-source transfer, the diffuse prior component, interaction modeling, target adaptation, and the intervention-acquisition objective (Table~\ref{tab:design-controls}). Additional experiments compare recurrent and feed-forward PPO, coverage rules, and planning variants (Table~\ref{tab:design-controls}). Because both investigators share the posterior estimator, their comparison concerns intervention selection rather than the inference model. Shuffled and contradictory source histories are used to assess negative transfer under prior mismatch.

\subsection{Independent counterfactual prediction}
An independent Qwen3-235B-A22B-Instruct-2507 predictor estimates target-reader correctness under context interventions. The predictor uses the fixed DeepInfra route, temperature 0, and a maximum of 400 output tokens. For each of 900 held-out questions and five target readers, it receives the question, accepted reference answer, four frozen passages, target-reader instructions, and one of five explanation conditions: none, borrowed ($B=0$), target-only ($B=32$), Bayesian-verified ($B=32$), or PPO-verified ($B=32$). Explanations use a deterministic rendering template; a single realized explanation is selected for each method and question. The predictor estimates correctness probabilities for the 11 subsets containing at most two passages. The six two-passage contexts are algebraically derivable from the full-context rate and reported deletion and interaction contrasts; the empty and four single-passage contexts require extrapolation. This audit assesses prediction of observable answer correctness, not model-internal computations. Because the accepted answer is available to the predictor and a single predictor model is evaluated, the results are conditional on that information setting.

Let $\widehat p_{q,t,S}^{\mathrm{pred}}$ be the predictor output, and let $\bar Y_{q,t,S}$ denote the mean of $R=16$ independent target-reference correctness observations. We calculate the noise-corrected Brier score as
\begin{equation}
\widehat{\mathrm{BS}}=\frac{1}{|\mathcal I|}\sum_{(q,t,S)\in\mathcal I}\left[\left(\widehat p_{q,t,S}^{\mathrm{pred}}-\bar Y_{q,t,S}\right)^2-\frac{\bar Y_{q,t,S}(1-\bar Y_{q,t,S})}{R-1}\right],
\label{eq:predictor-brier}
\end{equation}
where $\mathcal I$ is the set of evaluated question--target--subset triples. For predictions fixed independently of the reference responses, the correction removes the reference-sampling contribution to the expected squared error. The registered fallback probability for an unparseable predictor response is 0.5. Prediction prompts and outputs are committed before opening the independent reference bank. Paired comparisons use a dataset-stratified question bootstrap with 10,000 replicates, with Holm adjustment for the three primary comparisons. A sensitivity analysis excludes the Qwen3 target reader to evaluate predictor--target overlap (Table~\ref{tab:simulatability-tests}).

\subsection{Statistical interpretation}
\begin{figure}[tbp]
\centering
\includegraphics[width=\textwidth]{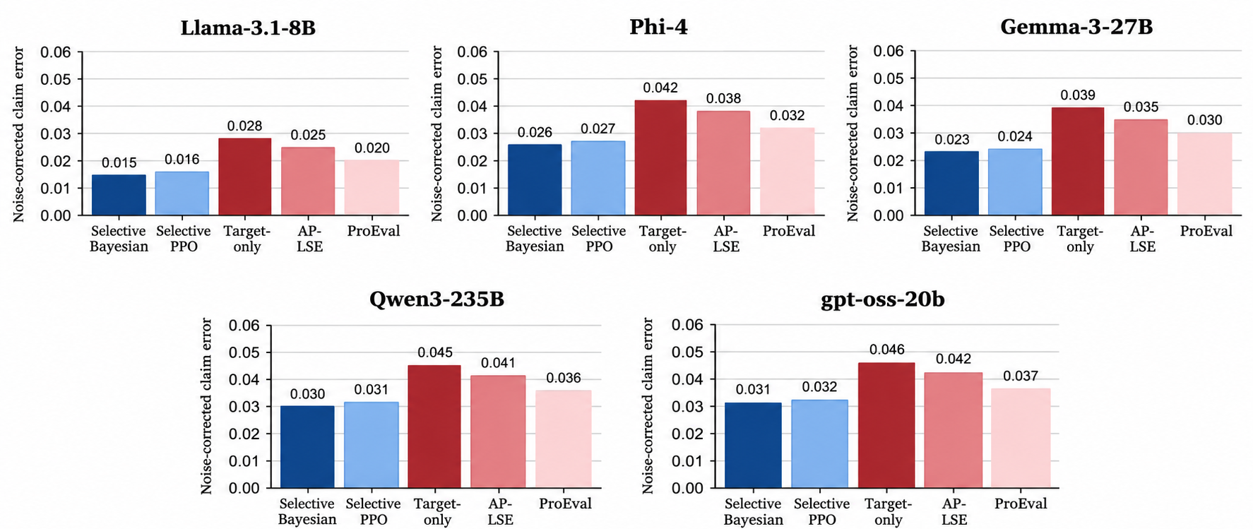}
\caption{Contrast-estimation error by target model for selective Bayesian transfer, selective PPO, target-only inference, AP-LSE and ProEval. Lower values are better.}
\label{fig:models}
\end{figure}

\begin{figure}[tbp]
\centering
\includegraphics[width=\textwidth]{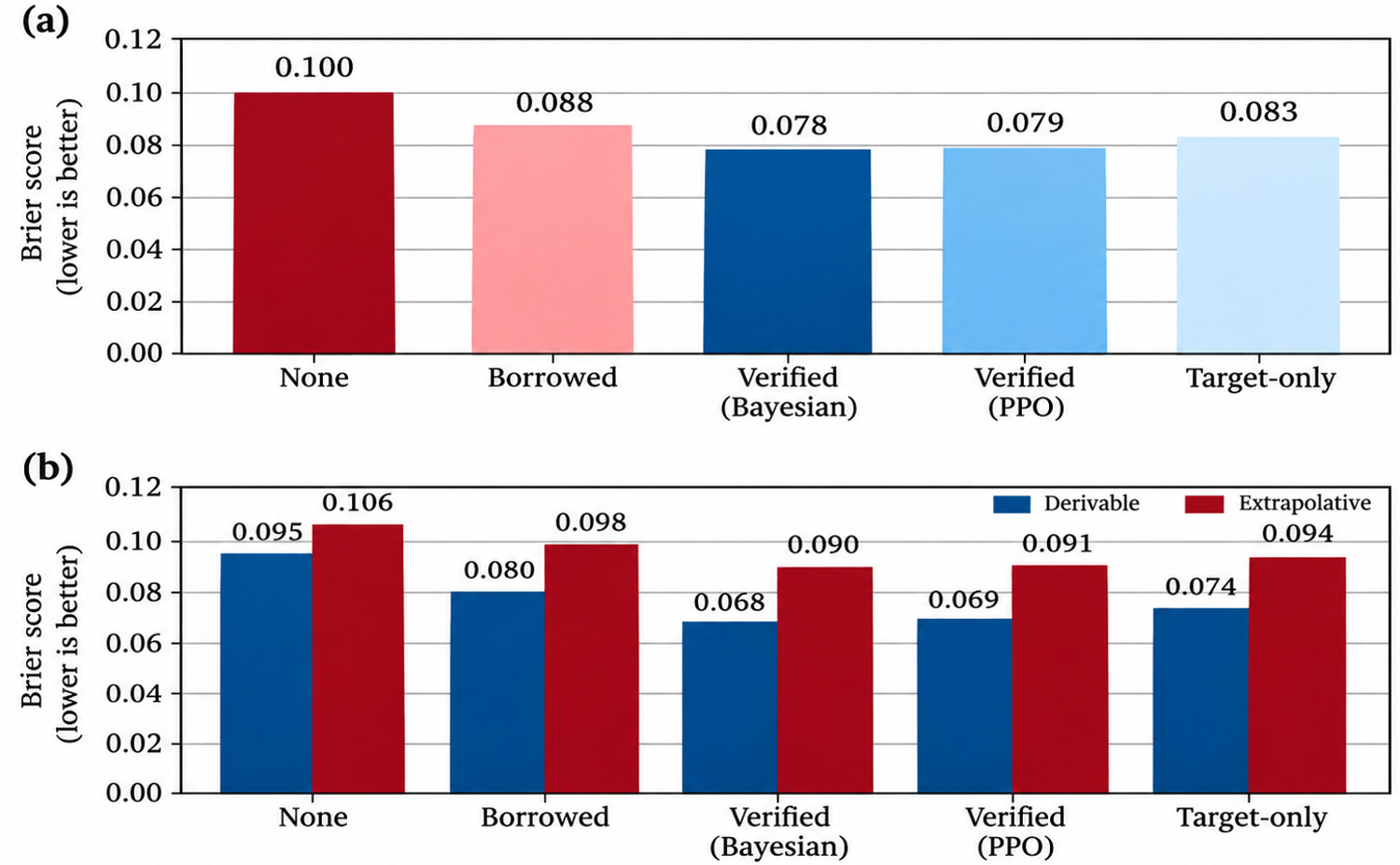}
\caption{Counterfactual simulatability measured using noise-corrected Brier scores: (a) overall scores under five explanation conditions and (b) scores for derivable and extrapolative evidence configurations. Lower scores are better.}
\label{fig:simulatability}
\end{figure}

\begin{figure}[tbp]
\centering
\includegraphics[width=\textwidth]{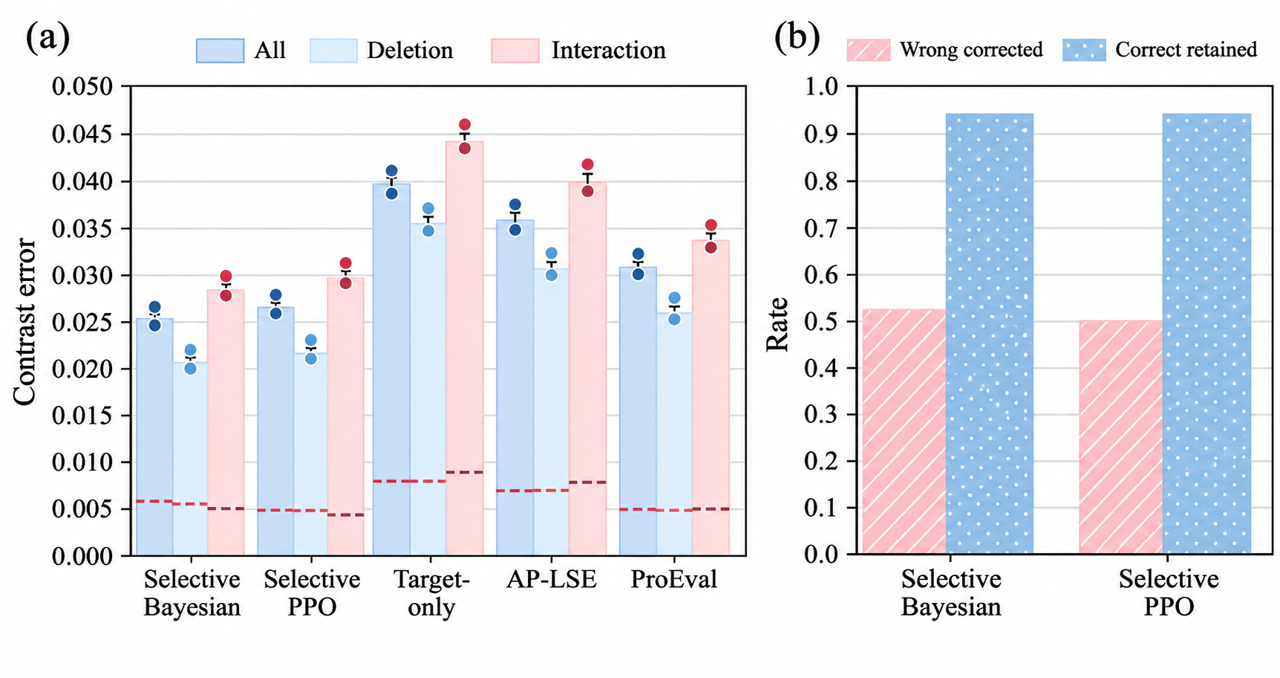}
\caption{Selective verification at 32 target observations: (a) pooled contrast-estimation error by claim type and method and (b) rates of correcting or withdrawing incorrect borrowed claims and retaining correct borrowed claims. Bars display reported point estimates only; no unreported uncertainty intervals are shown.}
\label{fig:verification}
\end{figure}

\begin{figure}[tbp]
\centering
\includegraphics[width=\textwidth]{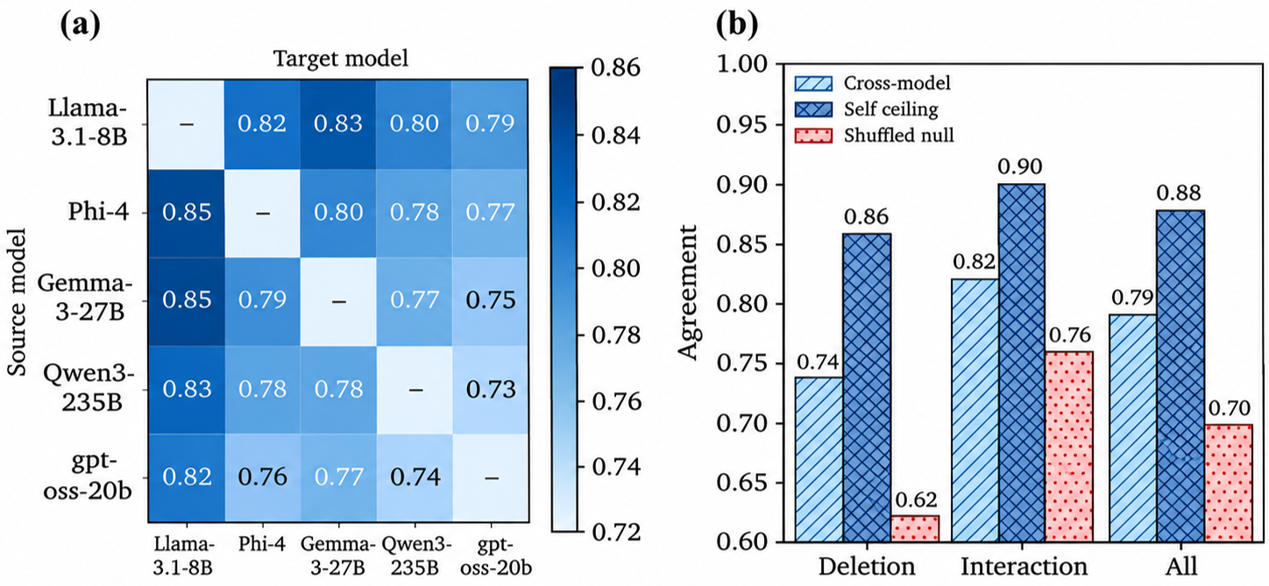}
\caption{Cross-model evidence-use agreement: (a) directed source--target expected region agreement and (b) pooled agreement for deletion, interaction and all claims, compared with same-model and shuffled-question controls.}
\label{fig:transfer}
\end{figure}

Inference uses paired question-level comparisons to preserve within-question dependence among claims and repeated responses. Uncertainty is clustered by question, and Holm adjustment is applied to the specified comparisons. Mean noise-corrected contrast error at $B=32$ is the principal endpoint; claim reliability, generalization, negative transfer, and counterfactual prediction are assessed separately. Figures~\ref{fig:models}--\ref{fig:transfer} and the accompanying tables provide the pooled, stratified, and control analyses.

\section{Results and analysis}
\label{sec:results}
Cross-reader agreement measures the transferability of deletion and interaction claims. Estimation, claim verification, ablations, source-mismatch experiments, and independent counterfactual prediction use the reported evaluation protocols and a primary target-response budget of 32.

\subsection{Cross-reader agreement and the reliability of borrowed claims}
Deletion and interaction claims agree more often across readers than under shuffled-question pairing, but less often than in same-reader comparisons (Fig.~\ref{fig:transfer}). For deletion claims, cross-reader agreement is 0.74, compared with 0.62 for shuffled pairing and 0.86 within a reader. Interaction agreement is 0.82, compared with 0.76 and 0.90, respectively. Because these aggregates include neutral regions, agreement does not by itself establish the reliability of material claims. Precision among initially confident borrowed claims is 0.81 for deletions and 0.72 for interactions (Table~\ref{tab:transfer}).

\begin{table}[tbp]
\centering
\caption{Transfer agreement and reliability of source-derived claims. Panel (a) summarizes initially confident borrowed claims; panel (b) compares expected region agreement across evaluation strata. The panels use distinct populations and measures.}
\label{tab:transfer}
\footnotesize
\begin{minipage}[t]{0.33\linewidth}
\centering
\textit{(a) Borrowed claim reliability}\par\smallskip
\setlength{\tabcolsep}{2pt}
\begin{tabular}{@{}lcc@{}}
\toprule
Claim & Precision & Reversal \\
\midrule
Deletion & 0.81 & 0.04 \\
Interaction & 0.72 & 0.07 \\
All claims & 0.77 & 0.05 \\
\bottomrule
\end{tabular}
\end{minipage}\hfill
\begin{minipage}[t]{0.65\linewidth}
\centering
\textit{(b) Agreement by evaluation stratum}\par\smallskip
\setlength{\tabcolsep}{2.5pt}
\begin{tabular}{@{}lcccc@{}}
\toprule
Stratum & Cross & Self & Shuffled & $\Delta$ \\
\midrule
HotpotQA & 0.75 & 0.87 & 0.64 & 0.12 \\
MuSiQue & 0.86 & 0.91 & 0.80 & 0.05 \\
2WikiMQA & 0.77 & 0.87 & 0.67 & 0.10 \\
Answerable context & 0.74 & 0.86 & 0.63 & 0.12 \\
Unanswerable context & 0.88 & 0.92 & 0.83 & 0.04 \\
\bottomrule
\end{tabular}
\end{minipage}
\par\vspace{2pt}
\raggedright\scriptsize $\Delta$ is self minus cross agreement.
\end{table}

Agreement differs across datasets and evidence-answerability strata. Cross-reader agreement is 0.74 for answerable contexts and 0.88 for unanswerable contexts, while the corresponding gaps from same-reader agreement are 0.12 and 0.04. Transfer direction also matters: Llama-to-Phi agreement is 0.82, compared with 0.85 for Phi-to-Llama (Fig.~\ref{fig:transfer}(a)). These differences motivate using source histories as uncertain prior information rather than copying source claims without target verification.

\subsection{Contrast estimation and paired comparisons}
With 32 target responses, Bayesian verification achieves a mean noise-corrected contrast error of 0.025, compared with 0.026 for PPO, 0.040 for target-only inference, 0.036 for adapted AP-LSE, and 0.031 for adapted ProEval (Fig.~\ref{fig:verification}(a)). The Bayesian-to-target-only reduction is 0.015, or 37.5\% of target-only error. For interaction claims, Bayesian and target-only errors are 0.028 and 0.044, respectively; these quantities measure response-surface contrasts rather than question-answering accuracy. Additional probabilistic diagnostics are reported in Table~\ref{tab:diagnostics}.

\begin{table}[tbp]
\centering
\caption{Probabilistic diagnostics and selective claim reliability at 32 target observations. Lower interaction error, Brier score, and conditional risk are better. Posterior and target-only certification are distinct uncertainty assessments; the latter uses the common target observations.}
\label{tab:diagnostics}
\footnotesize
\setlength{\tabcolsep}{3.5pt}
\begin{tabularx}{\linewidth}{@{}>{\raggedright\arraybackslash}Xccccc@{}}
\toprule
& \multicolumn{2}{c}{Selective transfer} & \multicolumn{3}{c}{Comparators} \\
\cmidrule(lr){2-3}\cmidrule(l){4-6}
Metric & Bayesian & PPO & Target-only & AP-LSE & ProEval \\
\midrule
\multicolumn{6}{@{}l}{\textit{Probabilistic diagnostics}} \\
Unscaled interaction error & 0.017 & 0.018 & 0.026 & 0.024 & 0.021 \\
Region Brier score & 0.220 & 0.225 & 0.300 & 0.280 & 0.250 \\
Interval mass & 0.82 & 0.81 & 0.88 & 0.74 & 0.70 \\
\addlinespace[2pt]
\multicolumn{6}{@{}l}{\textit{Claim reporting and certification}} \\
Posterior material-claim rate & 0.38 & 0.37 & 0.21 & 0.29 & 0.41 \\
Target-only certification rate & 0.09 & 0.09 & 0.09 & 0.09 & 0.09 \\
Coverage ($\kappa=0.80$) & 0.34 & 0.33 & 0.17 & 0.24 & 0.36 \\
Conditional risk ($\kappa=0.80$) & 0.11 & 0.11 & 0.14 & 0.16 & 0.17 \\
Coverage ($\kappa=0.90$) & 0.25 & 0.24 & 0.11 & 0.16 & 0.27 \\
Conditional risk ($\kappa=0.90$) & 0.07 & 0.07 & 0.09 & 0.11 & 0.12 \\
\bottomrule
\end{tabularx}
\end{table}

\begin{table}[tbp]
\centering
\caption{Paired primary comparisons with bootstrap uncertainty and multiplicity-adjusted inference. Negative differences favor the first method.}
\label{tab:comparisons}

\setlength{\tabcolsep}{2pt}

\begin{tabularx}{\linewidth}{@{}>{\raggedright\arraybackslash}Xcccc@{}}
\toprule
Comparison & Difference & 95\% CI & \shortstack{Family-wise\\CI} & $p_{\mathrm{Holm}}$ \\
\midrule
Bayesian vs Target-only & $-$0.0150 & [$-$0.0185, $-$0.0116] & [$-$0.0198, $-$0.0103] & $<$0.001 \\
Bayesian vs AP-LSE & $-$0.0110 & [$-$0.0142, $-$0.0079] & [$-$0.0154, $-$0.0067] & $<$0.001 \\
Bayesian vs ProEval & $-$0.0060 & [$-$0.0086, $-$0.0035] & [$-$0.0096, $-$0.0025] & $<$0.001 \\
PPO vs Target-only & $-$0.0140 & [$-$0.0176, $-$0.0105] & [$-$0.0189, $-$0.0092] & $<$0.001 \\
PPO vs AP-LSE & $-$0.0100 & [$-$0.0133, $-$0.0068] & [$-$0.0145, $-$0.0056] & $<$0.001 \\
PPO vs ProEval & $-$0.0050 & [$-$0.0077, $-$0.0024] & [$-$0.0087, $-$0.0013] & 0.002 \\
PPO vs Bayesian & +0.0010 & [$-$0.0004, +0.0024] & [$-$0.0009, +0.0029] & 0.31 \\
\bottomrule
\end{tabularx}
\end{table}

Paired question-level analysis supports lower error for both selective-transfer investigators than for target-only inference and the two adapted baselines (Table~\ref{tab:comparisons}). For Bayesian versus target-only inference, the paired difference is $-0.0150$ (95\% CI $[-0.0185,-0.0116]$; Holm-adjusted $p<0.001$). In contrast, the PPO-minus-Bayesian difference is $+0.0010$ (95\% CI $[-0.0004,0.0024]$; Holm-adjusted $p=0.31$). The reported comparison does not establish an estimation advantage for PPO over Bayesian acquisition.

\subsection{Transfer components and intervention-selection controls}
The combined ablation table separates Bayesian transfer components from PPO policy variants while using common error metrics (Table~\ref{tab:design-controls}). Replacing the coefficient-wise Bayesian prior with a whole-source prior increases overall error from 0.025 to 0.031. Removing target adaptation yields 0.033, omitting the diffuse prior component yields 0.028, and imposing an additive response surface yields 0.029. The latter variant has interaction error 0.034, compared with 0.028 for the full model. These contrasts are compatible with benefits from partial borrowing, target adaptation, and explicit interaction modeling under the evaluated protocol.

\begin{table}[tbp]
\centering
\caption{Transfer and acquisition controls at 32 target observations. All columns report noise-corrected contrast error. Shared baselines are listed once within each investigator family; lower error is better.}
\label{tab:design-controls}
\footnotesize
\setlength{\tabcolsep}{4pt}
\begin{tabularx}{\linewidth}{@{}l>{\raggedright\arraybackslash}Xccc@{}}
\toprule
Investigator & Configuration & All & Deletion & Interaction \\
\midrule
\multicolumn{5}{@{}l}{\textit{Bayesian acquisition and transfer components}} \\
Bayesian & Full coefficient-wise transfer & 0.025 & 0.020 & 0.028 \\
& Whole-source prior & 0.031 & 0.026 & 0.034 \\
& No diffuse prior component & 0.028 & 0.022 & 0.032 \\
& Additive response surface only & 0.029 & 0.021 & 0.034 \\
& No target adaptation & 0.033 & 0.027 & 0.037 \\
& Unconditional source mean & 0.032 & 0.026 & 0.036 \\
& Generic information gain & 0.027 & 0.022 & 0.030 \\
& Contrast-MSE acquisition & 0.026 & 0.021 & 0.029 \\
& No coverage queries & 0.026 & 0.021 & 0.029 \\
& Balanced allocation & 0.028 & 0.023 & 0.031 \\
\addlinespace[2pt]
\multicolumn{5}{@{}l}{\textit{PPO policy controls}} \\
PPO & Full recurrent policy & 0.026 & 0.021 & 0.029 \\
& Target-only prior & 0.037 & 0.031 & 0.041 \\
& Whole-source prior & 0.030 & 0.025 & 0.033 \\
& No coverage queries & 0.027 & 0.022 & 0.030 \\
& Feed-forward policy & 0.027 & 0.022 & 0.031 \\
\addlinespace[2pt]
\multicolumn{5}{@{}l}{\textit{Additional intervention-selection controls}} \\
Planning & Two-step lookahead & 0.025 & 0.020 & 0.028 \\
& Policy-guided planning & 0.025 & 0.020 & 0.028 \\
Adaptation & Step-DAD & 0.026 & 0.021 & 0.029 \\
\bottomrule
\end{tabularx}
\end{table}

Acquisition choices also affect estimation. Bayesian claim-directed selection has error 0.025, compared with 0.027 for generic information gain and 0.028 for balanced allocation. Recurrent PPO obtains 0.026; removing recurrence or coverage increases it to 0.027, while a target-only prior produces 0.037. Two-step lookahead and policy-guided planning each yield 0.025 at the reported precision. These rounded comparisons characterize the tested configurations but do not establish statistically significant differences for every ablation or an advantage for longer-horizon planning.

\subsection{Claim verification and selective reliability}
Verification is evaluated at the level of individual transferred claims, rather than only by average contrast error. Bayesian updating overturns or withdraws 52\% of initially incorrect confident borrowed claims and changes 6\% of initially correct ones, retaining 94\% of the latter (Fig.~\ref{fig:verification}(b)). PPO corrects or withdraws 50\% of incorrect claims and changes 6\% of correct claims. The denominators differ: the 52\% correction rate leaves 48\% of initially incorrect confident claims unchanged at the tested budget.

Claim-reporting thresholds further determine the trade-off between coverage and conditional risk (Table~\ref{tab:diagnostics}). At $\kappa=0.80$, Bayesian verification reports 34\% of claims at conditional risk 0.11, compared with 17\% coverage and risk 0.14 for target-only inference. Raising $\kappa$ to 0.90 reduces Bayesian coverage to 25\% and risk to 0.07. ProEval reports 36\% coverage at $\kappa=0.80$, but its conditional risk is 0.17. Source-informed posterior confidence must not be conflated with target-only interval certification, which is based on a different uncertainty calculation.

\subsection{Variation across target readers and datasets}
Bayesian contrast error is lower than target-only error in all 15 target-reader--dataset combinations (Fig.~\ref{fig:models}; Table~\ref{tab:per-model-dataset}). It ranges from 0.010 for Llama-3.1-8B on MuSiQue to 0.038 for gpt-oss-20b on HotpotQA. The reported stratification does not isolate the individual contributions of response stochasticity, evidence adequacy, or the prevalence of material effects. Full-context exact-match rates are included as a contextual diagnostic in Appendix Table~\ref{tab:baserates}, while the inferential comparison is based on pooled paired observations.

\begin{table}[tbp]
\centering
\caption{Noise-corrected contrast error for every target model and multi-hop dataset at 32 target observations. Lower values are better.}
\label{tab:per-model-dataset}

\setlength{\tabcolsep}{2pt}

\begin{tabular}{@{}llccccc@{}}
\toprule
Target & Dataset & Bayesian & PPO & Target-only & AP-LSE & ProEval \\
\midrule
Llama-3.1-8B & HotpotQA & 0.019 & 0.020 & 0.034 & 0.031 & 0.025 \\
Llama-3.1-8B & MuSiQue & 0.010 & 0.010 & 0.019 & 0.017 & 0.013 \\
Llama-3.1-8B & 2WikiMQA & 0.017 & 0.018 & 0.031 & 0.027 & 0.023 \\
\midrule
Phi-4 & HotpotQA & 0.032 & 0.033 & 0.051 & 0.048 & 0.039 \\
Phi-4 & MuSiQue & 0.017 & 0.018 & 0.028 & 0.025 & 0.021 \\
Phi-4 & 2WikiMQA & 0.029 & 0.030 & 0.046 & 0.041 & 0.036 \\
\midrule
Gemma-3-27B & HotpotQA & 0.029 & 0.030 & 0.048 & 0.044 & 0.037 \\
Gemma-3-27B & MuSiQue & 0.015 & 0.016 & 0.026 & 0.023 & 0.019 \\
Gemma-3-27B & 2WikiMQA & 0.026 & 0.027 & 0.043 & 0.038 & 0.034 \\
\midrule
Qwen3-235B & HotpotQA & 0.037 & 0.038 & 0.055 & 0.051 & 0.044 \\
Qwen3-235B & MuSiQue & 0.019 & 0.020 & 0.030 & 0.027 & 0.023 \\
Qwen3-235B & 2WikiMQA & 0.034 & 0.035 & 0.049 & 0.044 & 0.041 \\
\midrule
gpt-oss-20b & HotpotQA & 0.038 & 0.039 & 0.056 & 0.053 & 0.045 \\
gpt-oss-20b & MuSiQue & 0.020 & 0.021 & 0.031 & 0.028 & 0.024 \\
gpt-oss-20b & 2WikiMQA & 0.035 & 0.036 & 0.051 & 0.046 & 0.042 \\
\bottomrule
\end{tabular}
\end{table}

At the claim level, Bayesian region-decision accuracy ranges from 0.81 to 0.88 across targets, compared with 0.80 to 0.87 for PPO (Table~\ref{tab:verification-model}). Correction-related rates also differ by reader. Because overturned-claim rates and final region accuracy have different denominators, neither can replace the other as a measure of explanation reliability.

\begin{table}[tbp]
\centering
\caption{Claim-region accuracy (Acc.) and overturned-claim rate (Rev.) by target reader at 32 target observations. The two measures have different interpretations and should not be read as complementary rates.}
\label{tab:verification-model}
\footnotesize
\setlength{\tabcolsep}{3.4pt}
\begin{tabularx}{\linewidth}{@{}>{\raggedright\arraybackslash}Xcccccc@{}}
\toprule
& \multicolumn{2}{c}{Bayesian} & \multicolumn{2}{c}{PPO} & \multicolumn{2}{c}{Target-only} \\
\cmidrule(lr){2-3}\cmidrule(lr){4-5}\cmidrule(l){6-7}
Target reader & Acc. & Rev. & Acc. & Rev. & Acc. & Rev. \\
\midrule
Llama-3.1-8B & 0.88 & 0.55 & 0.87 & 0.53 & 0.83 & 0.60 \\
Phi-4 & 0.84 & 0.52 & 0.83 & 0.50 & 0.78 & 0.58 \\
Gemma-3-27B & 0.85 & 0.53 & 0.84 & 0.51 & 0.79 & 0.59 \\
Qwen3-235B & 0.82 & 0.50 & 0.81 & 0.48 & 0.76 & 0.56 \\
gpt-oss-20b & 0.81 & 0.49 & 0.80 & 0.47 & 0.75 & 0.55 \\
\bottomrule
\end{tabularx}
\end{table}

\subsection{Generalization beyond same-question source histories}
Removing same-question source histories increases Bayesian contrast error from 0.025 to 0.032 and PPO error from 0.026 to 0.034 (Table~\ref{tab:generalisation}). Both remain below the target-only error of 0.040. When each dataset is excluded from training in turn, Bayesian error ranges from 0.026 to 0.027. These tests examine transfer under restricted source information and held-out training datasets; they do not evaluate distribution shifts caused by changing retrieval systems or model endpoints.

\begin{table}[tbp]
\centering
\caption{Generalization under unavailable same-question source histories and held-out training datasets. Entries are noise-corrected contrast errors; lower values indicate better estimation.}
\label{tab:generalisation}

\setlength{\tabcolsep}{2.5pt}

\begin{tabular}{@{}>{\raggedright\arraybackslash}p{.29\linewidth}ccccc@{}}
\toprule
Method & Warm & \shortstack{New\\question} & \shortstack{Hold out\\HotpotQA} & \shortstack{Hold out\\MuSiQue} & \shortstack{Hold out\\2WikiMQA} \\
\midrule
Selective transfer, Bayesian & 0.025 & 0.032 & 0.027 & 0.026 & 0.027 \\
Selective transfer, PPO & 0.026 & 0.034 & 0.029 & 0.027 & 0.028 \\
Target-only & 0.040 & 0.040 & 0.040 & 0.040 & 0.040 \\
AP-LSE adaptation & 0.036 & 0.037 & 0.036 & 0.036 & 0.036 \\
ProEval adaptation & 0.031 & 0.036 & 0.033 & 0.032 & 0.033 \\
\bottomrule
\end{tabular}
\end{table}

\subsection{Independent counterfactual prediction}
Verified explanations are also evaluated by whether an independent predictor can estimate target-reader correctness under context interventions. The pooled Brier score is 0.100 without an explanation, 0.088 with a borrowed explanation, 0.083 with target-only inference, 0.078 after Bayesian verification, and 0.079 after PPO verification (Fig.~\ref{fig:simulatability}; Table~\ref{tab:counterfactual-summary}). Bayesian verification improves the score by 0.022 relative to no explanation (95\% CI [0.016, 0.028]) and by 0.010 relative to borrowing (95\% CI [0.005, 0.015]). Its 0.005 advantage over target-only explanations does not meet the adjusted significance criterion ($p=0.052$; Appendix Table~\ref{tab:simulatability-tests}).

\begin{table}[tbp]
\centering
\caption{Independent counterfactual-prediction Brier scores by explanation condition. The numerical-surface row reports a distinct numerical-information condition; no no-explanation result was reported for that condition. The remaining rows use the explanation-rendering protocol. Lower scores are better.}
\label{tab:counterfactual-summary}
\footnotesize
\setlength{\tabcolsep}{3.7pt}
\begin{tabularx}{\linewidth}{@{}>{\raggedright\arraybackslash}Xccccc@{}}
\toprule
Condition / stratum & None & Borrowed & Bayesian & PPO & Target-only \\
\midrule
\multicolumn{6}{@{}l}{\textit{Pooled and numerical-surface results}} \\
Pooled (rendered) & 0.100 & 0.088 & 0.078 & 0.079 & 0.083 \\
Pooled (numerical surface) & 0.09 & 0.071 & 0.060 & 0.061 & 0.066 \\
\addlinespace[2pt]
\multicolumn{6}{@{}l}{\textit{By target reader}} \\
Llama-3.1-8B & 0.072 & 0.061 & 0.055 & 0.056 & 0.059 \\
Phi-4 & 0.104 & 0.091 & 0.080 & 0.081 & 0.086 \\
Gemma-3-27B & 0.099 & 0.086 & 0.076 & 0.077 & 0.081 \\
Qwen3-235B & 0.110 & 0.098 & 0.086 & 0.087 & 0.092 \\
gpt-oss-20b & 0.115 & 0.104 & 0.093 & 0.094 & 0.097 \\
\addlinespace[2pt]
\multicolumn{6}{@{}l}{\textit{By dataset}} \\
HotpotQA & 0.118 & 0.104 & 0.092 & 0.093 & 0.098 \\
MuSiQue & 0.068 & 0.060 & 0.054 & 0.055 & 0.057 \\
2WikiMQA & 0.114 & 0.100 & 0.088 & 0.089 & 0.094 \\
\bottomrule
\end{tabularx}
\end{table}

The Bayesian-verified condition has lower scores than borrowing and no explanation in each reported reader and dataset stratum. In the numerical-surface control, where the evaluation uses numerical estimates rather than rendered explanations, Bayesian verification has error 0.060 versus 0.071 for borrowing. Figure~\ref{fig:simulatability}(b) separates evidence configurations whose outcomes are derivable algebraically from the stated contrasts from those requiring extrapolation. Predictive gains therefore concern observable intervention behavior; they do not establish access to the reader's internal computations.

\subsection{Negative transfer under misleading source histories}
Source information can increase error when its relation to the target is misspecified (Table~\ref{tab:harm}). Relative to target-only inference, Bayesian selective transfer has excess error $-0.0150$ with correctly aligned histories, but $+0.0012$ with shuffled histories and $+0.0031$ with contradictory histories. Under contradictory sources, the whole-source prior has excess error $+0.0240$, compared with $+0.0031$ for coefficient-wise Bayesian transfer. The diffuse component limits sensitivity to mismatched sources without eliminating negative transfer.

\begin{table}[tbp]
\centering
\caption{Paired excess contrast error relative to target-only inference and the percentage of questions harmed by transfer, under correct, shuffled, and contradictory source histories. Positive excess error denotes negative transfer.}
\label{tab:harm}
\footnotesize
\setlength{\tabcolsep}{3pt}
\begin{tabularx}{\linewidth}{@{}>{\raggedright\arraybackslash}Xcccccc@{}}
\toprule
& \multicolumn{2}{c}{Correct} & \multicolumn{2}{c}{Shuffled} & \multicolumn{2}{c}{Contradictory} \\
\cmidrule(lr){2-3}\cmidrule(lr){4-5}\cmidrule(l){6-7}
Method & Excess & Harmed & Excess & Harmed & Excess & Harmed \\
\midrule
Bayesian selective transfer & $-$0.0150 & 18\% & +0.0012 & 41\% & +0.0031 & 47\% \\
PPO selective transfer & $-$0.0140 & 19\% & +0.0015 & 42\% & +0.0036 & 48\% \\
AP-LSE adaptation & $-$0.0040 & 33\% & +0.0008 & 40\% & +0.0015 & 43\% \\
ProEval adaptation & $-$0.0090 & 26\% & +0.0060 & 55\% & +0.0170 & 68\% \\
Whole-source prior & $-$0.0090 & 27\% & +0.0100 & 60\% & +0.0240 & 74\% \\
No diffuse component & $-$0.0120 & 22\% & +0.0045 & 50\% & +0.0110 & 61\% \\
Unconditional source mean & $-$0.0080 & 29\% & +0.0035 & 49\% & +0.0090 & 58\% \\
\bottomrule
\end{tabularx}
\end{table}

\section{Discussion}
\label{sec:discussion}
Cross-reader explanation transfer requires both a source-to-target estimate and a decision about whether an individual borrowed claim remains supported. The reported results assess these distinct requirements through region agreement, contrast error, selective claim decisions, and source-mismatch controls. Their interpretation is limited to correctness under the specified passage interventions.

Context attribution methods estimate evidence dependence through intervention or approximation \citep{cohenwang2024contextcite,liu2025attribot}, while interaction estimators characterize non-additive evidence effects \citep{kang2025spex,butler2025proxyspex,jain2026cuer}. Cross-reader agreement exceeding shuffled-question agreement supports the use of source histories as informative prior data, but lower precision among initially confident borrowed interaction claims shows that agreement alone is insufficient for individual claim reporting. The proposed verifier addresses that distinction by updating the probability of each claim region using target outcomes rather than assuming source effects are preserved.

The coefficient-wise ablation and source-mismatch tests identify the observed contributions of partial borrowing and target adaptation. Whole-source transfer and removal of target adaptation increase contrast error; the diffuse component reduces, but does not eliminate, excess error under contradictory sources. These effects are specific to the defined four-passage response surface and the tested readers. AP-LSE and ProEval are task-specific adaptations, so their results do not constitute a comparison with those methods on their original objectives \citep{ngo2025aplse,huang2026proeval}.

Selective reporting exposes a separate reliability constraint. The verified Bayesian posterior changes a majority of initially incorrect confident borrowed claims while preserving most initially correct claims, but a substantial fraction of incorrect claims remain unresolved at 32 responses. Increasing the posterior confidence threshold reduces the fraction reported and its conditional risk. Source-informed posterior probabilities and target-only Clopper--Pearson certification represent different evidential standards and should not be interpreted as interchangeable guarantees.

Claim-directed acquisition has lower reported error than generic information gain and balanced allocation in the primary experiment, consistent with decision-focused experimental design \citep{rainforth2024modern,filstroff2024targeted}. The recurrent PPO policy does not establish a statistically supported improvement over Bayesian selection. The comparison consequently supports selective acquisition under the evaluated budget, not a claimed benefit of reinforcement learning or longer-horizon planning \citep{foster2021dad,blau2022optimizing,hedman2025stepdad}.

The Qwen3-based counterfactual-prediction audit assesses a downstream use of explanations and show lower Brier scores for verified than borrowed explanations. They do not establish access to internal reader mechanisms \citep{wallat2025correctness,chen2024simulatability}, and the advantage over target-only explanations does not meet the adjusted significance criterion. The predictor receives the accepted answer, and a separate non-Qwen target analysis assesses predictor--target overlap; neither design tests generalization to other predictor families.

The study fixes four passages, their order, a stationary decoding configuration, and exact-match correctness. Enumerating all contexts scales as $2^K$ in the number of passages. The evaluation covers three multi-hop datasets and five readers; contradictory sources still induce negative transfer and 48\% of initially incorrect confident claims remain unchanged after verification at the primary budget. Changes to the retrieval pipeline, larger passage sets, and explicit source-rejection rules require separate evaluation.

\section{Conclusion}
\label{sec:conclusion}
We formulated cross-reader evidence-use explanation transfer as selective verification of four passage-deletion effects and six pairwise interactions. Coefficient-wise source borrowing and Bayesian target-response acquisition reduce reported noise-corrected contrast error from 0.040 for target-only inference to 0.025 at 32 target responses. The verifier also supports claim-level withdrawal and reversal, although some incorrect borrowed claims persist and contradictory sources cause negative transfer. Recurrent PPO does not establish a significant improvement over Bayesian acquisition. The reported counterfactual-prediction scores favor verified explanations over borrowed explanations, although the advantage over target-only explanations does not meet the adjusted significance criterion.

\appendix
\renewcommand{\thetable}{A\arabic{table}}
\renewcommand{\theHtable}{appendix.\arabic{table}}
\setcounter{table}{0}
\section{Additional empirical comparisons}
\label{sec:appendix}
The appendix provides stratified contrast errors, paired counterfactual-prediction comparisons, and full-context correctness rates to supplement the main evaluation. These values retain the reported numerical precision and the same target-reader and dataset definitions used in the main text.

\begin{table}[tbp]
\centering
\caption{Additional stratified noise-corrected contrast errors. Dataset rows report error across all claim types; target-reader rows report interaction-claim error only. Lower values are better.}
\label{tab:per-dataset}
\footnotesize
\setlength{\tabcolsep}{4pt}
\begin{tabularx}{\linewidth}{@{}>{\raggedright\arraybackslash}Xccccc@{}}
\toprule
Stratum & Bayesian & PPO & Target-only & AP-LSE & ProEval \\
\midrule
\multicolumn{6}{@{}l}{\textit{Overall contrast error by dataset}} \\
HotpotQA & 0.031 & 0.032 & 0.049 & 0.045 & 0.038 \\
MuSiQue & 0.016 & 0.017 & 0.027 & 0.024 & 0.020 \\
2WikiMQA & 0.028 & 0.029 & 0.044 & 0.039 & 0.035 \\
\addlinespace[2pt]
\multicolumn{6}{@{}l}{\textit{Interaction-claim error by target reader}} \\
Llama-3.1-8B & 0.017 & 0.018 & 0.031 & 0.028 & 0.022 \\
Phi-4 & 0.029 & 0.030 & 0.046 & 0.042 & 0.035 \\
Gemma-3-27B & 0.026 & 0.027 & 0.043 & 0.039 & 0.033 \\
Qwen3-235B & 0.033 & 0.034 & 0.050 & 0.045 & 0.040 \\
gpt-oss-20b & 0.034 & 0.035 & 0.051 & 0.046 & 0.041 \\
\bottomrule
\end{tabularx}
\end{table}

\begin{table}[tbp]
\centering
\caption{Paired simulatability comparisons with reported intervals and adjusted significance levels.}
\label{tab:simulatability-tests}

\setlength{\tabcolsep}{3pt}

\begin{tabularx}{\linewidth}{@{}>{\raggedright\arraybackslash}Xcc@{}}
\toprule
Comparison & Gain [95\% CI] & $p_{\mathrm{Holm}}$ \\
\midrule
Bayesian verified vs none & 0.022 [0.016, 0.028] & $<$0.001 \\
Bayesian verified vs Borrowed ($B=0$) & 0.010 [0.005, 0.015] & $<$0.001 \\
Bayesian verified vs Target-only ($B=32$) & 0.005 [0.000, 0.010] & 0.052 \\
Excluding Qwen3 target: Bayesian verified vs none & 0.021 [0.014, 0.028] & $<$0.001 \\
Excluding Qwen3 target: Bayesian verified vs Borrowed ($B=0$) & 0.009 [0.003, 0.015] & 0.006 \\
Excluding Qwen3 target: Bayesian verified vs Target-only ($B=32$) & 0.004 [$-$0.001, 0.010] & 0.11 \\
\bottomrule
\end{tabularx}
\end{table}

\begin{table}[tbp]
\centering
\caption{Full-context exact match by target and dataset, used as a contextual diagnostic.}
\label{tab:baserates}

\setlength{\tabcolsep}{4pt}
\begin{tabular}{@{}lcccc@{}}
\toprule
Endpoint & HotpotQA & MuSiQue & 2WikiMQA & Mean \\
\midrule
Llama-3.1-8B & 0.36 & 0.11 & 0.21 & 0.23 \\
Phi-4 & 0.45 & 0.16 & 0.33 & 0.31 \\
Gemma-3-27B & 0.44 & 0.18 & 0.26 & 0.29 \\
Qwen3-235B & 0.45 & 0.17 & 0.38 & 0.33 \\
gpt-oss-20b & 0.43 & 0.24 & 0.42 & 0.36 \\
\midrule
Mean & 0.43 & 0.17 & 0.32 & 0.31 \\
\bottomrule
\end{tabular}
\end{table}

\begingroup
\footnotesize
\setlength{\itemsep}{0pt}

\endgroup

\end{document}